\documentclass[a4paper,twoside]{article}

\usepackage{epsfig}
\usepackage{subcaption}
\usepackage{calc}
\usepackage{amssymb}
\usepackage{amstext}
\usepackage{amsmath}
\usepackage{amsthm}
\usepackage{multicol}
\usepackage{pslatex}
\usepackage{apalike}
\usepackage{algorithm2e}
\usepackage[bottom]{footmisc}
\usepackage{SCITEPRESS}     
\usepackage{booktabs}
\usepackage[acronym,nonumberlist]{glossaries}
\newacronym{AD}{AD}{Autonomous Driving}
\newacronym{SLR}{SLR}{Systematic Literature Review}
\newacronym{CP}{CP}{Collaborative Perception}
\newacronym{CD}{CD}{Cooperative Driving}
\newacronym{ETSI}{ETSI}{European Telecommunications Standards Institute}
\newacronym{PDU}{PDU}{Packet Data Unit}
\newacronym{ITS}{ITS}{Intelligent Transportation System}
\newacronym{C-ITS}{C-ITS}{Cooperative-Intelligent Transportation Systems}
\newacronym{CPM}{CPM}{Cooperative Perception Message}
\newacronym{MCM}{MCM}{Maneuver Coordination Message}
\newacronym{DSRC}{DSRC}{Dedicated Short-Range Communication}
\newacronym{C-V2X}{C-V2X}{Cellular Vehicle-to-Everything}
\newacronym{LSTM}{LSTM}{Long Short-Term Memory}
\newacronym{V2X}{V2X}{Vehicle-to-Everything}
\newacronym{V2V}{V2V}{Vehicle-to-Vehicle}
\newacronym{V2I}{V2I}{Vehicle-to-Infrastructure}
\newacronym{IoV}{IoV}{Internet of Vehicles}
\newacronym{VRUs}{VRUs}{Vulnerable Road Users}
\newacronym{FOV}{FOV}{Field of View}
\newacronym{ROIs}{ROIs}{regions of interest}

\newacronym{AVs}{AVs}{Autonomous Vehicles}
\newacronym{CAVs}{CAVs}{Connected Autonomous Vehicles}
\newacronym{CAV}{CAV}{Connected Autonomous Vehicle}
\newacronym{BEV}{BEV}{Bird’s Eye View}
\newacronym{GNN}{GNN}{Graph Neural Network}

\newacronym{COD}{COD}{Collaborative Object Detection}
\newacronym{COT}{COT}{Collaborative Object Tracking}
\newacronym{CMP}{CMP}{Collaborative Motion Prediction}
\newacronym{CSS}{CSS}{Collaborative Semantic Segmentation}
\newacronym{CLD}{CLD}{Collaborative Lane Detection}

\newacronym{PandP}{P\&P}{Joint Perception and Prediction}
\newacronym{Co-PandP}{Co-P\&P}{Collaborative Joint Perception and Prediction}

\newacronym{HD Map}{HD Map}{High Definition Map}
\newacronym{GNSS}{GNSS}{Global Navigation Satellite System}
\newacronym{GPS}{GPS}{Global Positioning System}
\newacronym{SLAM}{SLAM}{Simultaneous Localization and Mapping}
\newacronym{CSC}{CSC}{Collaborative Scene Completion}

\begin{document}

\title{The Components of Collaborative Joint Perception and Prediction - A Conceptual Framework}

\author{\authorname{Lei Wan\sup{1}\sup{2}\orcidAuthor{0009-0007-4470-9088}, Hannan Ejaz Keen\sup{1}\orcidAuthor{0009-0001-6217-9427} and Alexey Vinel\sup{2}\orcidAuthor{0000-0003-4894-4134}}
\affiliation{\sup{1}XITASO GmbH, Augsburg, Germany}
\affiliation{\sup{2} Karlsruhe Institut of Technology, Karlsruhe, Germany}
\email{\{lei.wan, hannan.keen\}@xitaso.com, alexey.vinel@kit.edu}
}


\abstract{\acrfull{CAVs} benefit from \acrfull{V2X} communication, which enables the exchange of sensor data to achieve \acrfull{CP}. To reduce cumulative errors in perception modules and mitigate the visual occlusion, this paper introduces a new task, \acrfull{Co-PandP}, and provides a conceptual framework for its implementation to improve motion prediction of surrounding objects, thereby enhancing vehicle awareness in complex traffic scenarios. The framework consists of two decoupled core modules, \acrfull{CSC} and \acrfull{PandP} module, which simplify practical deployment and enhance scalability. Additionally, we outline the challenges in Co-P\&P and discuss future directions for this research area.}

\onecolumn \maketitle \normalsize \setcounter{footnote}{0} \vfill

\section{\uppercase{Introduction}} \label{sec:introduction}

\gls{AD}  technology is essential for advancing intelligent transportation systems, contributing to improved road safety, enhanced traffic efficiency, energy conservation, and reduced carbon emissions. A key component of the \gls{AD} framework is perception, which involves detecting dynamic objects and interpreting the static environment. The perception module encompasses various tasks, including object detection, tracking, motion prediction, and semantic segmentation. Traditionally, these tasks are implemented in a modular format, forming the basis for downstream functions like planning and control \cite{10.1007/978-3-031-32606-6_23,10.1007/978-3-030-19648-6_56}. Advancements in artificial intelligence and sensor fusion have significantly improved vehicle perception capabilities. However, single-vehicle perception still faces challenges, particularly with visual occlusion, which can pose safety risks and lead to accidents. \gls{V2X} technology offers a promising approach to address these limitations by enabling the sharing data with other vehicles or infrastructure, effectively enhancing perception and mitigating occlusion issues.

With \gls{V2X} communication, \gls{CAVs} can achieve \gls{CP} by integrating data from multiple sources. Initial research on \gls{CP} began within the communication field, focusing on standardizing V2X message types and optimizing communication efficiency. Recently, \gls{CP} research has expanded into computer vision and robotics, where the emphasis is shifting from sharing standardized messages, such as \gls{CPM} containing detected objects, to share raw sensor data or neural features. For example, Chen et al. \cite{FcooperFeatureBased-2019-chend} propose a feature-based \gls{CP} approach that transmits and combines LiDAR features across vehicles, enhancing perception performance within bandwidth constraints. Similarly, Hu et al. \cite{CollaborationHelpsCamera-2023-hub} present a camera-based \gls{CP} method that integrates visual \gls{BEV} features from multiple agents, providing a more comprehensive view of dynamic objects.

\begin{figure}[t]
\centering
\includegraphics[width=\linewidth]{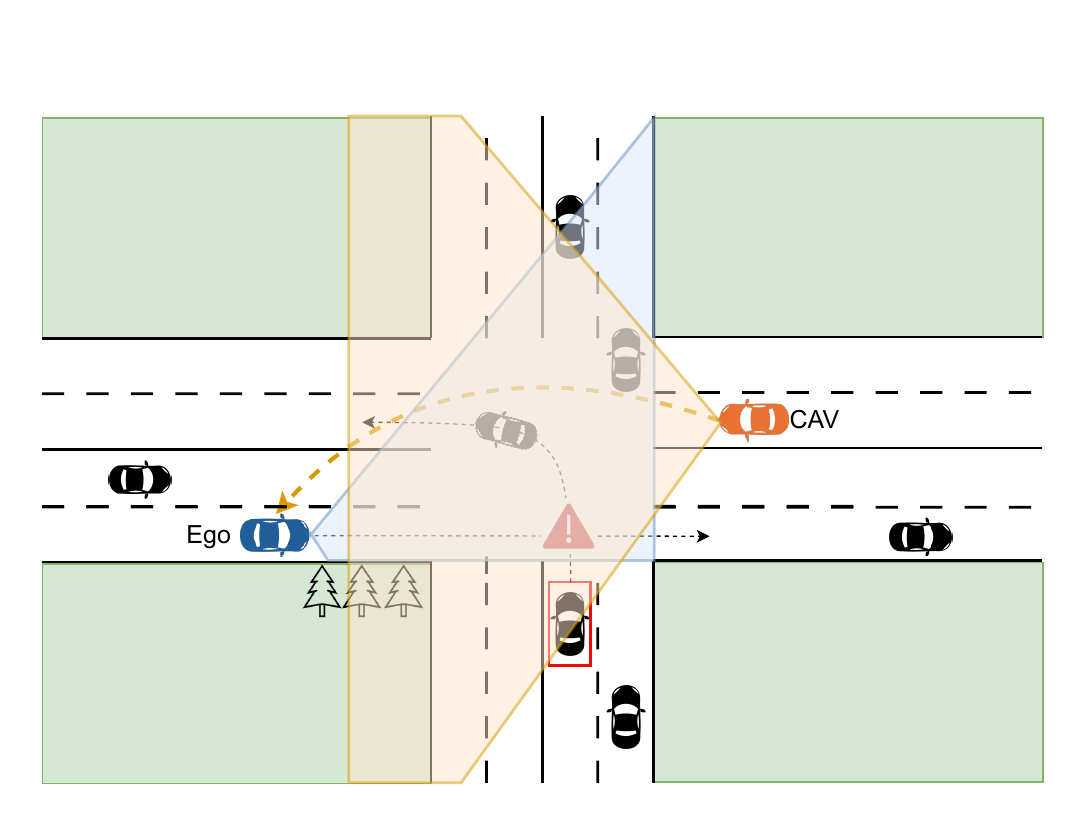}
\caption{Schematic diagram of Collaborative Perception. The diagram illustrates a scenario at an intersection where two \gls{CAVs} collaborate to enhance perception. The ego vehicle (blue) has a limited field of view due to occlusions, such as trees and buildings, which block its line of sight to a vehicle turning left. A collaborating vehicle (orange) positioned across the intersection shares its sensor data, expanding the ego vehicle's awareness. The shaded areas represent the \gls{FOV} for each vehicle}
\label{fig:cp}
\end{figure}

The use of collaborative methodologies extend beyond object detection to enhance other perception tasks. For instance, Liu et al. \cite{LiDARSemanticSegmentation-2023-liua} introduce a collaborative semantic segmentation framework utilizing intermediate collaboration, achieving superior results compared to single-vehicle methods. In motion prediction, Wang et al. \cite{V2VNetVehicletoVehicleCommunication-2020-wangb} demonstrate how collaboration enhances the precision of predicted trajectories.Additionally some perception tasks are handled within multi-task pipelines, such as the V2XFormer model by Wang et al. \cite{DeepAccidentMotionAccident-2024-wanga}, which simultaneously outputs detection, motion prediction, and semantic segmentation results.While multi-task approaches benefit from resource savings by sharing a common backbone, they often overlook temporal information across sensor frames, which is essential for tracking and motion prediction.

An emerging trend is the development of differentiable frameworks that seamlessly integrate various perception tasks within a unified model, enabling end-to-end training. For example, Liang et al. \cite{liang2020pnpnet} propose an end-to-end \gls{PandP} framework for single vehicles equipped with LiDAR. Similarly, Gu \cite{gu2023vip3d} introduces a camera-only end-to-end pipeline for \gls{PandP}, utilizing visual features to achieve both detection and motion prediction. These approaches highlight the potential of end-to-end learning to address bottlenecks in traditional perception pipelines, where cumulative errors across stages can degrade performance. With end-to-end learning, cumulative noise is mitigated, and motion prediction benefits significantly from the integration of fine-grained contextual information. Nonetheless, current research in \gls{PandP} still encounters challenges, particularly with visual occlusion, which significantly impacts prediction accuracy for obscured targets.

To overcome this issue, we propose the \gls{Co-PandP} framework, which incorporates \gls{V2X} collaboration. Our framework is based on the premise that \gls{CP} complements ego-vehicle perception, making it adaptable to scenarios with or without \gls{V2X} support. Additionally, to simplify deployment and enhance scalability, our framework decouples the training of the collaboration module from perception tasks. Inspired by recent work \cite{MultiRobotSceneCompletion--lib,CoreCooperativeReconstruction-2023-wanga}, our approach uses collaborative scene completion to address visual occlusion. Consequently, the \gls{Co-PandP} framework comprises two core modules: the \acrfull{CSC} module and the \acrfull{PandP} module.

In addition to developing \gls{CP} approaches, establishing effective evaluation methodologies is crucial for advancing \gls{CP} research. Current studies largely adopt evaluation methods designed for single-vehicle perception. Notably, only one study \cite{UMCUnifiedBandwidthefficient-2023-wangb} has introduced an evaluation focused on invisible objects, highlighting \gls{CP}'s potential to address visual occlusion. New evaluation methods are also required to assess the performance of \gls{Co-PandP}.

The main contributions of this paper are as follows:
\begin{itemize}
    \item We introduce a conceptual framework for \gls{Co-PandP}, designed to address cumulative errors inherent in modular designs and mitigate visual occlusion challenges.
    \item We present a re-formulation of evaluation methods in \gls{CP} and propose an evaluation approach tailored for \gls{Co-PandP} that aligns with the motivation of \gls{V2X} collaboration.
    \item We outline the challenges surrounding \gls{Co-PandP}, and suggest future work to further enhance this framework.
\end{itemize}

The structure of the paper is as follows: Section \ref{sec:system} details the system design, while Section \ref{sec:evaluation} introduces the evaluation method for \gls{Co-PandP}. Sections \ref{sec:challenges} discusses real-world challenges regarding practical deployment. Finally, Section \ref{sec:conclusion} concludes with a summary and outlook.


\begin{figure*}[ht]
\centering
\includegraphics[width=\linewidth]{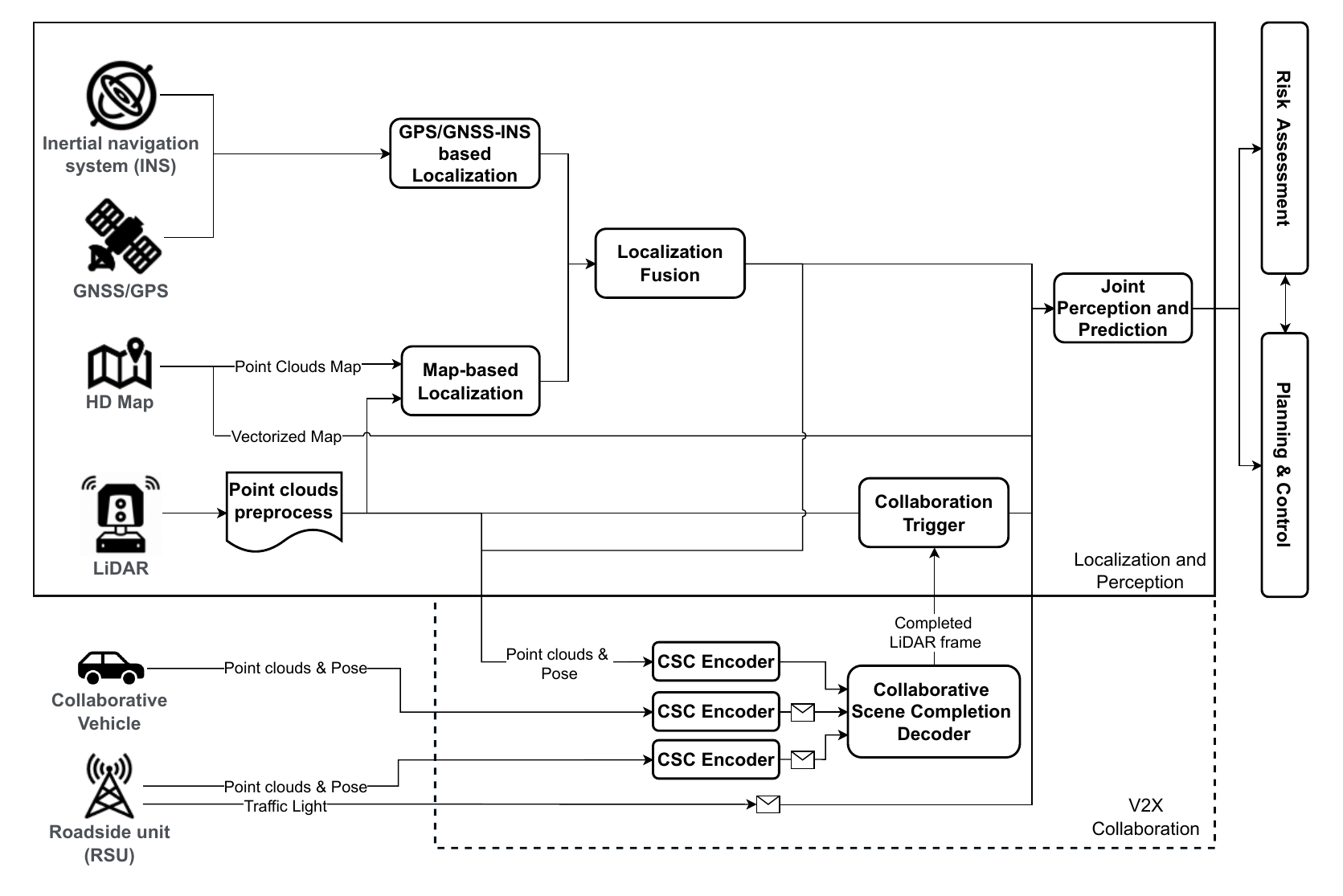}
\caption{Schematic diagram of Collaborative Joint Perception and Prediction.The system combines GPS/GNSS-INS and map-based localization for precise positioning. Sensing data (point clouds, poses) from collaborative vehicle and roadside unit are processed and shared with the ego vehicle, enabling \gls{CSC} to provide a comprehensive LiDAR frame. To optimize bandwidth usage, intermediate features are shared for scene completion instead of raw point clouds are shared via \gls{V2X}. The collaboration trigger manages \gls{CSC} activation. The \gls{PandP} module integrates localization, map, and LiDAR data to jointly enhance perception and prediction, which feeds into the risk assessment and planning and control modules for real-time decision-making.}
\label{fig:cp_framework}
\end{figure*}

\section{\uppercase{Details of the System}} \label{sec:system}
Figure \ref{fig:cp_framework} presents the conceptual framework, encompassing sensors, localization, \gls{HD Map}, communication, \gls{PandP}, collaborative scene completion. This section provides an overview of each core component of the framework along with their corresponding approaches.


\subsection{Sensors}
To capture a 3D view of the environment, various sensor types can be employed, including LiDAR, radar, and different types of cameras such as RGB and infrared camera. In our conceptual framework, LiDAR serves as the primary sensor due to its high precision in 3D measurements, significantly enhancing the 3D perception capabilities of \gls{AVs}.

LiDAR sensors vary in their scanning patterns, typically classified into spinning and oscillating types \cite{triess2021survey}. Spinning LiDAR uses a regular scanning pattern that provides an even distribution of points across a $360^\circ$ \gls{FOV}. In contrast, oscillating LiDAR follows a snake-like pattern, creating a denser yet uneven point distribution within a constrained \gls{FOV}. Each type offers distinct characteristics that can lead to a domain gap in perception models due to differences in data representation. Addressing this domain gap across different LiDAR sensor is essential for \gls{CP} systems. 


\subsection{Localization}
Beyond environmental perception, precise self-localization is essential for \gls{CP}. Accurate localization allows for data fusion across dynamic agents by establishing a consistent coordinate system to align all sensory data. Thus, the effectiveness of \gls{CP} depends significantly on the localization accuracy of \gls{CAVs}.

Traditionally, vehicles rely on \gls{GNSS} or \gls{GPS} to determine their position using trilateration. However, \gls{GNSS}-based methods face challenges like Non-Line-of-Sight and multipath propagation, often resulting in errors exceeding 3 meters, which undermines reliability and safety in \gls{AD} \cite{OCHIENG2002171}. \gls{HD Map} can mitigate localization errors, achieving centimeter-level accuracy \cite{chalvatzaras2022survey}. These maps are created through extensive data collection runs, often using LiDAR to construct a detailed point cloud layer. For precise vehicle positioning on \gls{HD Map}, both \gls{GNSS} and LiDAR sensors are used, providing a precise positioning approach.

\subsection{\gls{HD Map}}
\gls{HD Map} serves not only for localization but also offer essential semantic information about the static environment. They include detailed road data such as lane boundaries, lane centerlines, road markings, traffic signs, poles, and traffic light locations. This information aids vehicles in interpreting traffic rules and understanding the surrounding environment, enhancing motion prediction accuracy. Xu et al. \cite{Xu2023TowardsMF} highlight the significant impact of map quality on motion prediction performance, showing that high-quality, curated \gls{HD Map} outperform systems relying on online mapping or operating without maps. In our framework, the map operates as an independent module that interfaces with the perception module. This design enables compatibility with various mapping solutions, supporting scalability to online mapping or even cost-efficient, mapless approaches.

\subsection{Communication}
\gls{V2X} communication technology forms a critical foundation for \gls{CP}. \gls{CAVs} and intelligent infrastructure use sensors to perceive the environment and then transmit this data through \gls{V2X} communication. Two primary technologies support V2X communications: \gls{DSRC} and cellular network technologies \cite{abboud2016interworking}.

\gls{DSRC} is a wireless technology designed for automotive and \gls{C-ITS} applications, allowing short-range information exchange between devices. It operates without additional network infrastructure and offers low latency, making it suitable for safety-critical applications \cite{5888501}. However, \gls{DSRC} has limitations, including a relatively short communication range and reduced scalability in scenarios with high vehicular density \cite{harding2014vehicle}. Cellular networks, on the other hand, offer a potential solution for \gls{C-ITS} by providing greater bandwidth. These capabilities ensure that sensor data, crucial for \gls{CP}, can be effectively transmitted across distributed entities. Yet, some \gls{C-V2X} modes depend on cellular infrastructure, meaning performance may degrade in areas far from base stations, impacting latency.

Given the limitations of using either \gls{V2X} technology alone, a hybrid approach that combines both \gls{DSRC} and cellular technologies is more promising, enabling novel \gls{DSRC}–cellular interworking schemes. In our framework, data such as traffic light information, which requires low bandwidth, is well-suited for \gls{DSRC}. Meanwhile, sensor data, with its higher bandwidth demands, is more effectively handled by \gls{C-V2X}.

\subsection{Joint Perception and Prediction}
\begin{figure}[t]
\centering
\includegraphics[width=\linewidth]{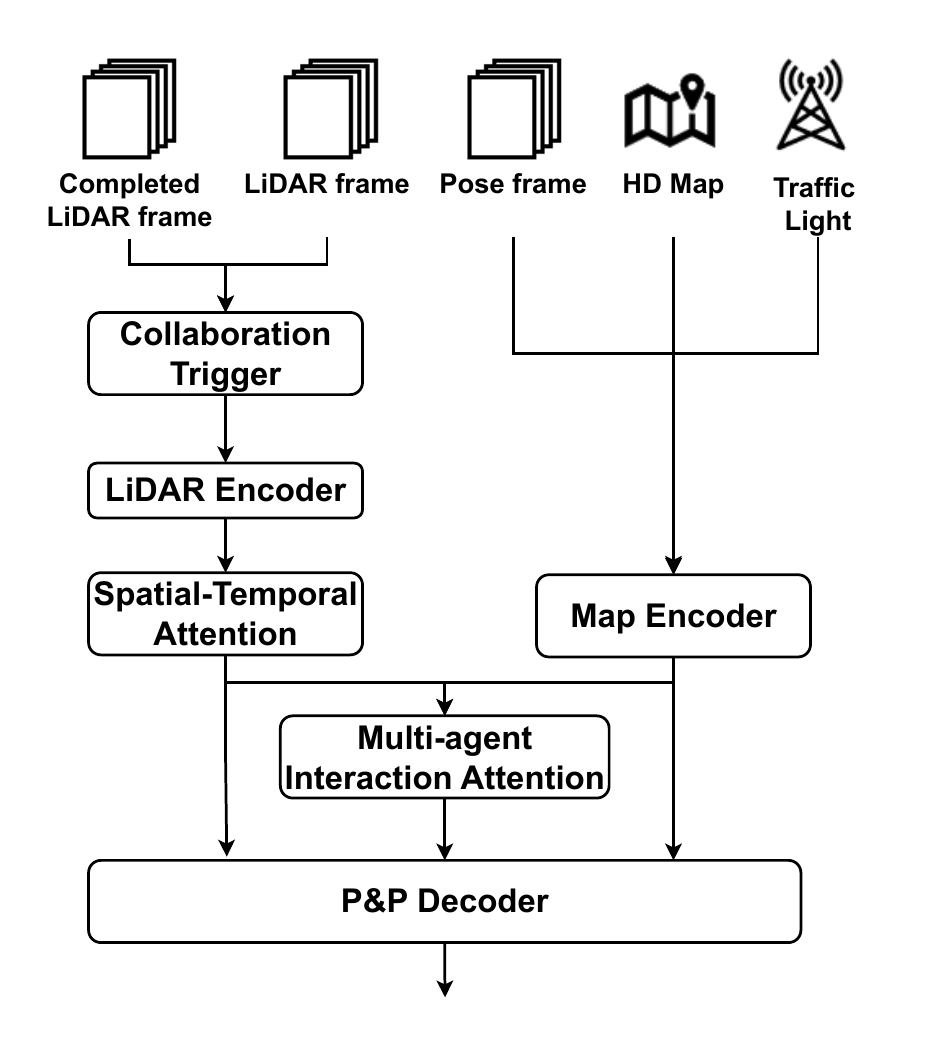}
\caption{Schematic diagram of \gls{PandP}.}
\label{fig:pandp_framework}
\end{figure}

The \gls{PandP} module forms the core of our framework. This module integrates data from LiDAR, vehicle pose, \gls{HD Map}, and traffic light information to generate detection results and forecast the trajectories of relevant agents, as depicted in Figure \ref{fig:pandp_framework}. The \gls{PandP} pipeline includes a LiDAR encoder, temporal encoder, map encoder, multi-agent interaction encoder, and \gls{PandP} decoder.

\begin{itemize}
    \item{{\bf{Lidar Encoder}}: To enable semantic understanding of the surrounding environment for \gls{AVs}, the LiDAR encoder extracts semantic features from point clouds. For example, VoxelNet \cite{zhou2018voxelnet} divides the point cloud into a 3D voxel grid, aggregates features within each voxel, and encodes these features. By applying voxel convolution, it captures 3D spatial features, which are then flattened along the z-axis and transformed into \gls{BEV} features,  enhanceing computational efficiency}
    \item{{\bf{Spatial-Temporal Attention}}:In addition to spatial features from the LiDAR encoder, temporal information across multiple frames is crucial for understanding temporal dynamics of the environment \cite{bharilya2024machine}. In our framework, the temporal encoder captures this temporal context from LiDAR \gls{BEV} features. For instance, a cross-attention mechanism \cite{vaswani2017attention} extracts context between frames, generating spatial-temporal features.}
    \item{{\bf{Map Encoder}}:Map and traffic light information are crucial for motion prediction \cite{ettinger2021large}. To interact with spatial-temporal features, the map and traffic light data are encoded as neural features. For instance, the \gls{HD Map} is transformed into the ego-vehicle’s coordinate system, centered on the ego-vehicle, and only map information within a defined surrounding area is used. Traffic light data are integrated into the map as environmental indicators and encoded as features.}
    \item{{\bf{Multi-Agent Interaction Attention}}: Modeling interactions among multiple agents is challenging \cite{bharilya2024machine}. This module combines agent and map features to ensure more precise modeling of these interactions. This block first computes interactions between the map and agent information, then calculates inter-agent interactions within \gls{ROIs}.}
    \item{{\bf{\gls{PandP} Decoder}}:The decoder integrates all relevant features to produce accurate perception and prediction outputs. Detection results are represented as a \gls{BEV} map comprising a map mask and object masks. Motion prediction is represented as a \gls{BEV} flow output, aligning well with downstream tasks such as planning and decision-making.}
\end{itemize}

\subsection{Collaboration Trigger}
While multi-agent collaboration provides significant benefits for \gls{CAVs}, it also demands substantial resources. Collaboration is often unnecessary when the ego vehicle has unobstructed visibility. To balance system effectiveness and efficiency, a collaboration trigger is needed to activate collaboration only at optimal times. Designing an effective collaboration trigger and identifying relevant decision factors remain underexplored areas of research \cite{huang2023v2x}. In our framework, we consider scenario occlusion levels, the confidence level of the ego vehicle’s perception, and communication conditions in developing this trigger metric. When the metric value exceeds a specified threshold, the system activates the collaboration module

\subsection{Collaborative Scene Completion} \label{sec:co_scene_complete}
Traditional \gls{CP} frameworks generally involve the sharing of neural features generated by deep learning-based \gls{CP} modules or the exchange of perception results among cooperative agents. However, this approach is task-specific, meaning that the shared data can only support particular perception tasks, leading to heterogeneity of systems that limits effective collaboration between diverse agents \cite{han2023collaborative}. Additionally, conventional methods often require joint model training across agents and, in some cases, re-training the whole-model for each perception task \cite{MultiRobotSceneCompletion--lib}. This joint training can be impractical and resource-intensive in real-world applications.

\begin{figure}[t]
\centering
\includegraphics[width=\linewidth]{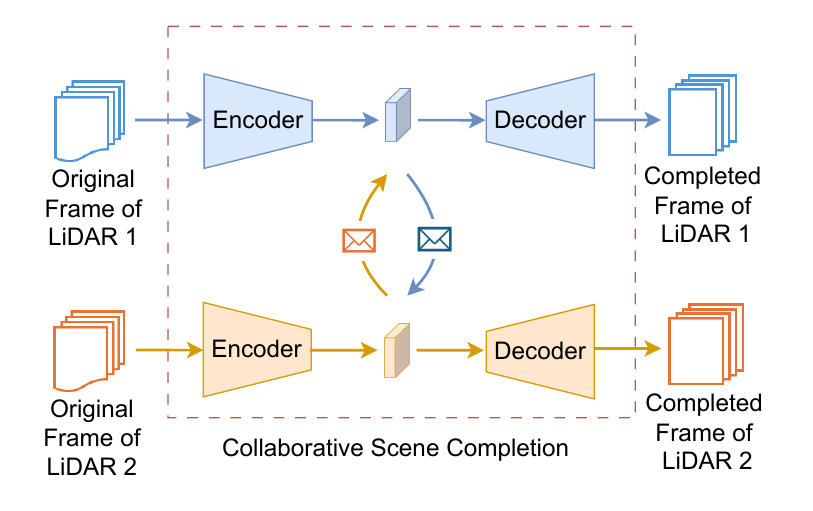}
\caption{Schematic diagram of \acrfull{CSC}.}
\label{fig:csc}
\end{figure}

In our framework, we decouple \gls{V2X} collaboration from the \gls{PandP} pipeline, allowing independent training of each component. \gls{V2X} collaboration is managed through task-agnostic collaborative scene completion, which benefits all downstream tasks without needing task-specific data transmission. By reconstructing a comprehensive scene using latent features and sharing these features across agents, this approach minimizes communication demands, as shown in Figure \ref{fig:csc}. The completed scene is then fed into the \gls{PandP} pipeline, as shown in Figure \ref{fig:pandp_framework}.

\section{\uppercase{Evaluation}} \label{sec:evaluation}
In addition to developing the \gls{Co-PandP} system,it is essential to establish effective evaluation methods to accurately access its performance. However, evaluating \gls{CP} presents unique challenges. Most existing research relies on methods adapted from single-vehicle perception, which fail to reflect \gls{CP}’s capacity to address visual occlusions \cite{UMCUnifiedBandwidthefficient-2023-wangb}. To overcome this limitation, new evaluation methods are required.

\begin{table}[t]
\centering
\caption{Summary of Metrics for Evaluating Joint Perception and Prediction}
\label{tab:metrics}
\begin{tabular}{c|p{5cm}}
\toprule
Metrics & Description   \\ \midrule
$minADE_k$ & the minimum over $k$ predictions of Average Distance Error: the average of point-wise L2 distances between the prediction and ground-truth forecasts   \\ \midrule
$minFDE_k$ & the minimum over $k$ predictions of Final Distance Error: the L2 distance at the final future time step  \\ \midrule
$MR_{k@x}$ & the MissRate: the ratio of forecasts having $minFDE_k > x=4m$  \\ \midrule
$mAP_f$ & the Mean Forecasting Average Precision: adapted from $mAP_{det}$, $mAP_f$ additionally penalizes trajectories that have correct first-frame detections but inaccurate forecasts ($minFDE_k < 4m$), but also trajectories with incorrect first-frame detections (center distance $<2m$)   \\ 
\bottomrule
\end{tabular}
\end{table}

\subsection{Evaluation method}
As \gls{CP} complements single-vehicle perception, primarily aiming to resolve visual occlusions, evaluation metrics should reflect precision on objects that are hidden from an individual vehicle’s view but visible from a collaborative perspective. For example, Wang et al. \cite{UMCUnifiedBandwidthefficient-2023-wangb} propose the Average Recall of Collaborative View (ARCV) metric, which measures recall for agents invisible from a single-vehicle perspective but detectable through collaboration. Our evaluation method follows this approach by categorizing objects into three groups: fully visible, partially visible, and fully invisible. We assess the perception performance of algorithms across these categories, first without collaboration and then with \gls{V2X} collaboration, to measure the improvement provided by \gls{CP} systems.

Apart from perception metrics, communication cost is also critical. In our evaluation method, we use average message size as an effective metric to assess the communication demands of collaborative perception methods \cite{marez2022bandwidth}.

Evaluating \gls{PandP} introduces additional challenges, especially in comparing traditional modular method, where detection, tracking, and prediction are conducted sequentially, and end-to-end methods, which directly process sensor data to generate perception and prediction results in a unified framework. Both approaches should receive the same detection and tracking inputs for the forecasting module. For instance, Xu et al. \cite{Xu2023TowardsMF} introduce a method to evaluate both traditional and end-to-end forecasting models, using the metrics summarized in Table \ref{tab:metrics}. A primary metric in their approach is Mean Forecasting Average Precision ($mAP_f$), inspired by detection $AP$ \cite{peri2022forecasting}. In our work, $mAP_f$ and $mAP_{det}$ are the principal metrics used to assess detection and forecasting performance across different object groups: fully visible, partially visible, and fully invisible.


\subsection{Evaluation in Simulation}
\gls{Co-PandP} is a complex multi-agent system influenced by factors such as localization error and communication constraints. To evaluate the robustness of this approach, ablation studies on key factors are essential. Simulation provides a practical solution, as it offers a fully controlled environment for testing. In our work, we use simulation to conduct various ablation studies to assess \gls{Co-PandP}’s performance under different conditions, including localization error, latency, and the number of \gls{CAVs}. This process ensures the scalability of our approach across diverse real-world scenarios. Future research will benefit from advanced co-simulators that support realistic communication and sensor data for even more comprehensive testing.

\subsection{Evaluation with Real-world Dataset}
Benchmarking perception algorithms on real-world datasets is a standard approach for evaluating and comparing methods, as real-world data offers a higher degree of realism. For \gls{Co-PandP} research, DAIR-V2X-Seq \cite{V2XSeqLargeScaleSequential-2023-yub} is a useful dataset, containing 7,500 cooperative frames with infrastructure and vehicle-side images and point clouds. However, \gls{PandP} relies heavily on machine learning, which requires large-scale dataset. The scale of DAIR-V2X-Seq remains limited for training larger ML models. To advance \gls{Co-PandP} research, creating a more extensive \gls{CP} dataset is crucial, and it is one of our primary goals for future work.

\section{\uppercase{Challenges}} \label{sec:challenges}
While \gls{Co-PandP} has significant potential to enhance vehicle awareness in dynamic traffic environments by reducing accumulated errors and addressing occlusion issues, its real-world implementation faces several challenges. This section outlines key challenges in deploying \gls{Co-PandP}.

\begin{itemize} 
\item {{\bf{Localization errors}}: Effective sensor data fusion requires aligning all data in a shared coordinate system, which depends on precise vehicle localization. However, \gls{GNSS}-based localization typically varies in accuracy from 1 to 3 meters, leading to potential misalignments that can significantly impair data fusion. Addressing these pose errors is essential for accurate collaborative scene completion in our framework.}
\item {{\bf{Asynchronous}}: Collaborative scene completion becomes more complex due to asynchronous observations from multiple agents. To accurately reconstruct a current scene frame, input from other perspectives is often necessary. However, these inputs are frequently asynchronous with the ego vehicle’s observations, causing inconsistencies in the positions of dynamic objects. Developing methods to handle asynchronous data effectively is critical for accurate scene completion.}
\item {{\bf{Domain Shift}}: In real-world traffic, vehicles from various manufacturers may be equipped with different types of LiDAR sensors, such as rotating and oscillating LiDARs. Variations in scan patterns lead to distinct data representations across sensors, introducing domain shifts that can disrupt the perception pipeline \cite{DIV2XLearningDomainInvariant-2023-xiangb,HPLViTUnifiedPerception-2023-liuc}. To prevent performance degradation, it is crucial to develop methods for completing the LiDAR scene using each sensor’s unique data representation.}
\item {{\bf{Dependency on Large-scale Labeled Dataset}}: The \gls{PandP} module employs a unified neural network without hand-crafted processing steps, such as Non-Maximum Suppression (NMS). This high degree of neural network reliance increases data demands during model training. Similar to end-to-end driving models, end-to-end \gls{PandP} models require large datasets. Reducing dependency on annotated data is essential to streamline \gls{PandP} deployment, presenting a critical area for further investigation.}
\end{itemize}

\section{\uppercase{Conclusion}}
\label{sec:conclusion}

In this paper, we introduced a conceptual framework for \gls{Co-PandP}, which comprises collaborative scene completion and \gls{PandP} module. By decoupling V2X collaboration from perception, the framework enables separate training and validation of the two modules, supporting scalable deployment in real-world settings. A significant challenge in collaborative scene completion is bridging the domain gap between different LiDAR sensors, which we propose to address using a unified intermediate representation format, similar to that used in 3D reconstruction. After revisiting evaluation methods in \gls{CP}, we emphasize that evaluating \gls{CP} performance on objects at different visibility level provides valuable insights, particularly for objects that are fully invisible from ego view but visible from collaborative perspective. This metric highlights \gls{CP}'s potential to address visual occlusion, which should be considered a primary motivation for \gls{CP}. Additionally, we discuss the challenges and open questions surrounding \gls{Co-PandP}.

This conceptual framework serves as a high-level architecture for \gls{Co-PandP}, with detailed implementation of each component to follow in future work. In addition to developing novel modules for collaborative scene completion and \gls{PandP}, creating a large-scale dataset is essential to advance this field. We plan to develop a large-scale dataset supporting a range of CP tasks, including detection, tracking, and motion prediction.

\bibliographystyle{apalike}
{\small
\bibliography{main}}

\end{document}